\documentclass[aps,prl,twocolumn,superscriptaddress,floatfix]{revtex4-2}

\usepackage{graphicx}
\usepackage{color}
\usepackage{xcolor}

\usepackage{amsmath}
\usepackage{amssymb}
\usepackage{amsfonts}
\usepackage{bm}  

\usepackage{booktabs}
\usepackage{multirow}

\usepackage{hyperref}
\hypersetup{
    colorlinks=true,
    linkcolor=blue,
    citecolor=blue,
    urlcolor=blue,
}

\begin{document}



\title{Grokking as Dimensional Phase Transition in Neural Networks}



\author{Ping Wang}
\affiliation{Institute of High Energy Physics, Chinese Academy of Science, 100049 Beijing, China}
\email{pwang@ihep.ac.cn}


\date{\today}


\begin{abstract}


Neural network grokking---the abrupt memorization-to-generalization 
transition---challenges our understanding of learning dynamics. Through 
finite-size scaling of gradient avalanche dynamics across eight model 
scales, we find that grokking is a \textit{dimensional phase transition}: 
effective dimensionality~$D$ crosses from sub-diffusive (subcritical, $D < 1$) to 
super-diffusive (supercritical, $D > 1$) at generalization onset, exhibiting 
self-organized criticality (SOC). Crucially, $D$ reflects \textbf{gradient 
field geometry}, not network architecture: synthetic i.i.d.\ Gaussian 
gradients maintain $D \approx 1$ regardless of graph topology, while real 
training exhibits dimensional excess from backpropagation correlations. 
The grokking-localized $D(t)$ crossing---robust across topologies---offers 
new insight into the trainability of overparameterized networks.

\end{abstract}

%
%
%


\maketitle



The training dynamics of deep neural networks remain poorly understood despite their 
remarkable empirical success. A striking example is ``grokking''~\cite{power2022grokking}: 
during training on algorithmic tasks, models exhibit an abrupt transition from 
memorization to generalization---training accuracy reaches near-perfect levels while 
test accuracy remains at chance, then suddenly jumps to perfect generalization. 
This sharp phase transition is puzzling: standard learning theory 
does not predict why a network that already fits the training data 
perfectly~\cite{zhang2021understanding} should later improve further 
on a test set, let alone why this improvement occurs abruptly 
rather than gradually. A common thread 
across proposed explanations is an \emph{abrupt learning transition}---a sudden 
reorganization of internal representations---whose underlying gradient-level 
mechanism remains uncharacterized.

Several explanations have been proposed, including circuit 
formation~\cite{nanda2023progress}, representation learning~\cite{liu2023grokking}, 
circuit efficiency~\cite{varma2023explaining}, and phase-transition-like training 
dynamics~\cite{rubin2024grokking}, but these remain qualitative. 
We address this by investigating quantitatively whether grokking behaves as a 
\emph{dimensional phase transition} governed by 
SOC~\cite{bak1987self,bak1988self}---a universal mechanism for phase transitions in 
complex systems ranging from earthquakes to brain 
networks~\cite{beggs2003neuronal,chialvo2010emergent}. 
Our key finding: \textbf{effective dimensionality reflects gradient field geometry}, 
not network architecture. Real training exhibits a \emph{dimensional phase 
transition}: the effective dimensionality~$D$---the FSS exponent in 
$s_{\max} \sim N^D$, measuring how avalanche extent scales with system size---extracted 
via finite-size scaling (FSS) of gradient avalanche dynamics across multiple 
model sizes, evolves from sub-diffusive ($D < 1$) to super-diffusive ($D > 1$) states, crossing 
the random-diffusion baseline ($D = 1$) during generalization;
synthetic i.i.d.\ Gaussian gradients maintain $D\approx 1$ invariant
to topology, confirming dimensional evolution reflects
backpropagation's correlations.

\begin{figure*}
\includegraphics[width=\textwidth]{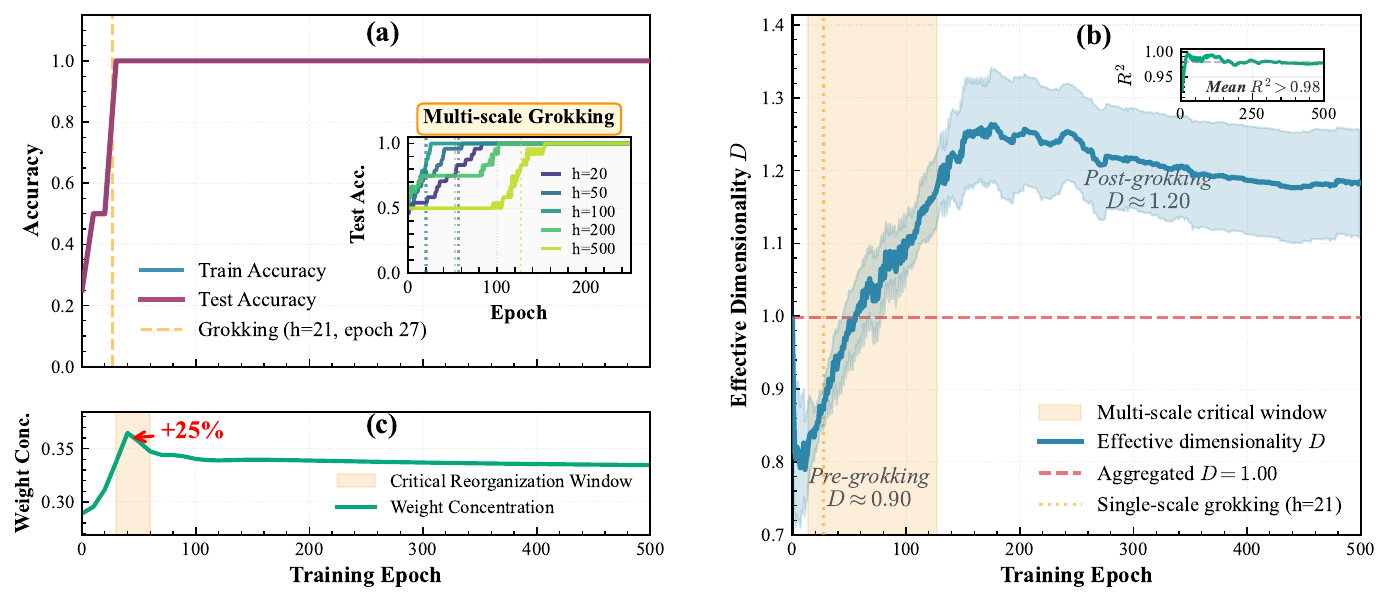}
\caption{\textbf{Grokking as transient SOC and dimensional phase transition.} 
\textbf{(a)} Training (blue) and evaluation (purple) accuracies for representative 
XOR case (h=21, N=85; train and evaluation share the same four patterns), 
showing a synchronized abrupt transition at epoch 27. Inset: 
multi-scale analysis across h=20--500 reveals scale-dependent grokking timing 
spanning epochs 12--134. \textbf{(b)} Time-resolved FSS analysis shows 
effective dimensionality $D$ evolves continuously during training. Yellow 
region: multi-scale grokking window. Orange line: single-scale grokking. 
Red line: time-averaged $D = 1.00 \pm 0.02$. Inset: FSS fit quality $R^2 > 0.98$. 
\textbf{(c)} Representative example: Weight concentration (Gini coefficient 
of $|\boldsymbol{\theta}|$; teal) exhibits transient peak coinciding with 
grokking. Multi-seed statistical validation (1000 seeds) described in text.}
\label{fig:overview}
\end{figure*}

\begin{figure}
\includegraphics[width=\columnwidth]{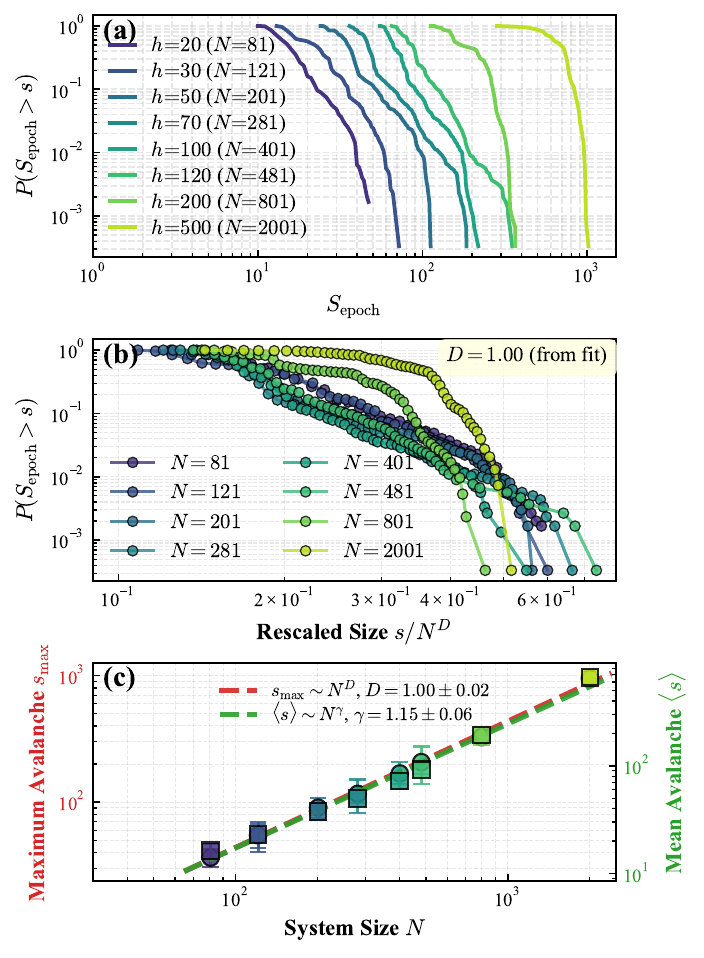}
\caption{\textbf{Finite-size scaling analysis of avalanche dynamics.}
\textbf{(a)} Complementary cumulative distributions (CCDF) of avalanche
sizes across eight model scales ($h = 20$--$500$), showing heavy-tailed,
scale-dependent behavior with systematic cutoff growth.
\textbf{(b)} X-only data collapse: plotting $P(>s)$ vs $s/N^D$ collapses
all scales toward a common curve using a single exponent~$D$,
validating the FSS exponent without additional fitting parameters.
\textbf{(c)} FSS of maximum ($s_{\max} \sim N^D$, left axis) and mean
($\langle s \rangle \sim N^\gamma$, right axis) avalanche sizes, yielding
$D = 1.00 \pm 0.02$ ($R^2 = 1.00$) and $\gamma = 1.15 \pm 0.06$
($R^2 = 0.99$) across eight scales.}
\label{fig:fss}
\end{figure}

\begin{figure}
\includegraphics[width=\columnwidth]{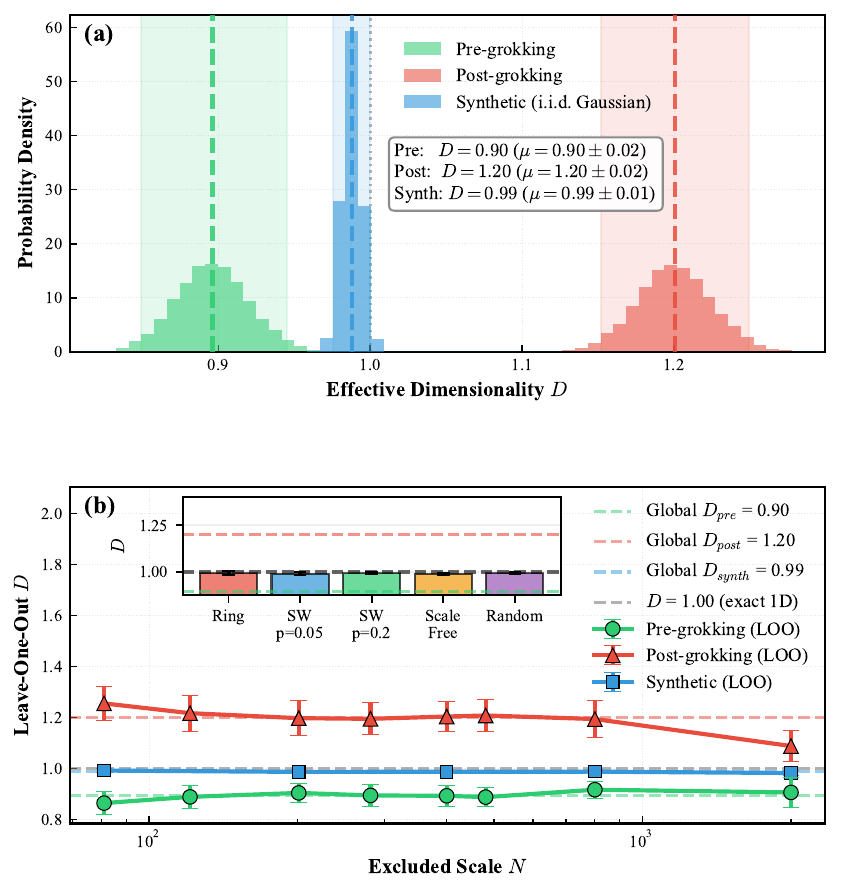}
\caption{\textbf{Gradient Geometry Determines Dimensionality.} 
\textbf{(a)} Bootstrap distributions (10,000 resamples) of the FSS exponent~$D$,
where each run is phase-split at its own grokking epoch:
pre-grokking real gradients (green, $D = 0.90 \pm 0.02$, 
sub-diffusive), post-grokking real gradients (red, $D = 1.20 \pm 0.02$, 
super-diffusive), and synthetic i.i.d.\ Gaussian gradients (blue, 
$D = 0.99 \pm 0.01$). Three non-overlapping peaks confirm statistically 
distinct scaling regimes.
\textbf{(b)} Leave-One-Out FSS analysis: removing any single scale preserves 
$D$, confirming scale invariance across $N = 81$--$2001$. Inset: 
Five network topologies collapse to $D \approx 0.99$ for synthetic 
gradients, demonstrating topology invariance.}
\label{fig:mechanism}
\end{figure}

We present evidence in three stages, with $D(t)$ as the unifying quantity:
(1) \textbf{Time-resolved evolution} (Fig.~\ref{fig:overview})---$D(t)$ evolves 
from sub-diffusive ($D \approx 0.90$) through the random-diffusion baseline 
$D \approx 1$ to super-diffusive ($D \approx 1.20$) during generalization, 
spanning a 30\% dynamic range;
(2) \textbf{Aggregate scaling analysis} (Fig.~\ref{fig:fss})---heavy-tailed,
scale-dependent distributions collapse across eight model scales with
$D \approx 1.0$ and $\gamma \approx 1.15$ ($R^2 > 0.99$);
(3) \textbf{Phase-resolved validation} (Fig.~\ref{fig:mechanism})---bootstrap 
analysis reveals two statistically distinct scaling regimes 
($D_{\mathrm{pre}} = 0.90$, $D_{\mathrm{post}} = 1.20$), demonstrating 
that grokking induces a crossover from sub-diffusive to super-diffusive 
cascade dynamics; topology invariance (coefficient of variation, 
CV $< 0.3\%$) confirms dimensionality reflects gradient field geometry, 
not network architecture.
Cross-task validation on modular arithmetic and ungrokked-run negative 
controls (both from companion study~\cite{wang2026companion}) further 
confirm criticality is grokking-specific.

Rigorous FSS requires systematic variation of system 
size---analogous to studying phase transitions via lattice sizes in Ising 
models~\cite{onsager1944crystal,goldenfeld1992lectures}. We use the XOR boolean 
function as a \emph{controlled} minimal testbed: its small dataset enables dense 
temporal sampling and precise grokking-epoch identification across eight model 
scales ($N = 81$--$2001$, spanning 1.4 decades). We note that XOR lacks a 
separate test split---training and evaluation use the same four patterns---so 
the transition we observe is an \emph{abrupt learning transition} in gradient 
geometry rather than canonical delayed generalization. This is a methodological 
feature, not a limitation: it isolates the gradient-level phase transition from 
behavioral confounds. A companion study~\cite{wang2026companion} independently
confirms the identical $D(t)$ signature in canonical grokking
(Transformer on ModAdd-59, 80/20 train/test split), establishing
that the gradient mechanism generalizes to the classical setting.

%
%
%


During backpropagation, gradients across different parameters acquire
correlations through shared loss landscape structure and the chain
rule~\cite{saxe2014exact,jacot2018neural}. If these correlations are
strong, a perturbation in one gradient component propagates to many
others---analogous to how the correlation length diverges at phase
transitions in spin systems~\cite{goldenfeld1992lectures}. To quantify
this correlation structure, we introduce the \emph{ Threshold-based 
Diffusion Update inspired by the Olami-Feder-Christensen earthquake
model}~\cite{ChristensenOlami1992OFC,olami1992self} (TDU-OFC) as an
in-line \emph{measurement probe}: real training gradients are injected
as initial conditions into a threshold-driven diffusion process, the
redistributed gradients feed into each parameter update, and we measure
how far perturbations cascade. This is analogous to tracer diffusion in
fractal media, where effective spatial dimension is extracted from
diffusion scaling through the system's own response to perturbations.
Although TDU-OFC introduces significant local modifications to gradient
geometry, a
companion study~\cite{wang2026companion} with shadow-probe controls
($\alpha_{\mathrm{train}} = 0$, diffusion excluded from training
entirely) confirms that the $D(t)$ crossing persists unchanged.
The macroscopic observable~$D$ is therefore insensitive to these local
perturbations: the observed transition reflects the underlying training
dynamics, not an artifact of the probe.

Standard stochastic gradient descent (SGD) updates parameters
independently: $\theta'_i = \theta_i - \eta \nabla_i L$. In TDU-OFC,
all $N$ trainable parameters (concatenated without layer distinction
into a single index array) are mapped onto a diffusion graph---here a
Barab\'asi-Albert (BA) scale-free network~\cite{barabasi1999emergence}
($m=2$, $\langle k \rangle = 4$), chosen for computational convenience---and
gradients exceeding a self-organizing threshold
$\tau = Q_{90}(|\nabla L|)$ (90th percentile, computed per epoch;
robust across $Q_{80}$--$Q_{95}$)
trigger diffusion to neighbors, generating
\emph{avalanches}---cascades of parameter updates. At each diffusion
step, for nodes $i$ with $|g_i| > \tau$, gradients redistribute via
\begin{equation}
g_i' = (1-\alpha)\,g_i, \quad
g_j' = g_j + \frac{\alpha\,g_i}{k_i} \;\;\text{for all } j\sim i,
\label{eq:diffusion}
\end{equation}
where $k_i$ is the degree (number of neighbors) of node $i$,
$\alpha = 0.3$ is diffusion strength (robust across
$\alpha = 0.1$--$0.5$, CV $< 0.4\%$), and $j \sim i$ denotes graph
neighbors. Degree normalization ensures quasi-conservative dynamics.
Iteration continues (max 20 steps; real training typically requires
$<$10) until no nodes exceed threshold. The avalanche size $s$ counts
total triggered updates---measuring how far gradient perturbations
propagate. Larger avalanches indicate more parameters are effectively
coupled; This implements the condensed matter paradigm of probing internal
correlations through \emph{relaxation response}: the avalanche is
the system's relaxation to above-threshold perturbations, and its
size quantifies the spatial extent of this relaxation. The mean avalanche size serves
as a generalized susceptibility~\cite{sethna2001crackling,pruessner2012self}
whose growth with system size signals criticality.

We extract effective dimensionality $D$---the FSS exponent in
$s_{\max} \sim N^D$ across system sizes $N$. For i.i.d.\ Gaussian gradients, mean total cascade size
$\langle S\rangle \sim N^{D_{\mathrm{synth}}}$ yields
$D_{\mathrm{synth}} = 0.99 \pm 0.01 \approx 1$
(distinct from the per-epoch peak-cascade statistic used for training data);
$\Delta D \equiv D - 1$ quantifies the excess for real training gradients.
Although TDU-OFC introduces substantial local
modifications to gradient geometry---the redistributed gradient vector
deviates $\sim\!30^\circ$ from the original in parameter space---$D$
remains topology-invariant, confirming it captures macroscopic
correlation structure rather than probe artifacts.

We study the XOR boolean function
(4 samples, 2 inputs $\to$ 1 output) via multilayer perceptrons
(Input[2] $\to$ Hidden[$h$] $\to$ Output[1], $N = 4h + 1$ parameters)
with binary cross-entropy loss and SGD (learning rate $\eta = 0.5$,
500 epochs). Hidden sizes $h \in \{20, 30, 50, 70, 100, 120, 200, 500\}$
with 51 gradient snapshots recorded at 10-epoch intervals are used, 
and 6 independent seeds per scale.

Our central finding is that grokking manifests as a \emph{dimensional
phase transition}: the effective dimensionality~$D$, extracted via
FSS of gradient avalanche dynamics, evolves
continuously during training and crosses the random-diffusion baseline
$D \approx 1$ at generalization onset. We present this evidence in three
stages---time-resolved evolution (Fig.~\ref{fig:overview}), aggregate
scaling analysis (Fig.~\ref{fig:fss}), and phase-resolved validation
(Fig.~\ref{fig:mechanism})---with $D(t)$ as the unifying quantity.

Figure~\ref{fig:overview}a shows a representative XOR trajectory
($h = 21$, $N = 85$): training and evaluation accuracies jump abruptly
at epoch~27. Multi-scale analysis (inset) reveals
scale-dependent grokking timing spanning epochs 12--134---a
\emph{scale-dependent reorganization process} that motivates
dimensional analysis via FSS.

Time-resolved FSS across eight model scales and six
seeds~\cite{onsager1944crystal,goldenfeld1992lectures} reveals systematic
evolution of the effective dimensionality $D$ (defined via
$s_{\max} \sim N^D$). Figure~\ref{fig:overview}b shows $D$ transitions
from $D \approx 0.90$ pre-grokking---\emph{below} the i.i.d.\ Gaussian
baseline ($D \approx 1.0$)---through continuous rise during the multi-scale
grokking window (yellow region), to $D \approx 1.20$ post-grokking,
representing a 30\% dynamic range. This gradual evolution reframes
grokking as a \textit{geometric phase transition} where the system
crosses from sub-diffusive gradient dynamics to super-diffusive
coordination. Concurrent with this dimensional transition, weight
concentration (Gini coefficient of $|\boldsymbol{\theta}|$;
Figure~\ref{fig:overview}c) exhibits a transient
+25\% peak lasting $\sim$50 epochs at the generalization transition,
providing an independent structural signature of reorganization.
Validated across 1000 seeds, peak timing synchronizes tightly with
grokking (within $\pm$10 epochs), distinguishing this as brief critical
reorganization rather than sustained criticality.

To test whether 
the observed criticality corresponds to self-organized criticality, we analyze 
avalanche size distributions using complementary cumulative distribution 
functions (CCDF) across eight model 
scales~\cite{sethna2001crackling}. 
Figure~\ref{fig:fss}a reveals heavy-tailed, scale-dependent distributions 
for all hidden sizes, with systematic cutoff growth 
$s_{\max} \sim N^D$ characteristic of finite-size SOC systems. The progressive 
rightward shift of the cutoff with increasing system size provides direct 
visual evidence for scale-invariant dynamics: larger systems sustain larger 
avalanches, a hallmark of criticality. Cross-task validation via ModAdd-59 
(a companion study~\cite{wang2026companion}) confirms similar heavy-tailed scaling 
over broader dynamic range ($\sim$120k parameters, $\sim$60$\times$ our largest 
XOR scale), supporting universality of the underlying SOC mechanism.
Strict MLE power-law fitting~\cite{clauset2009power} of individual CCDFs is
unreliable at our per-scale dynamic range ($<$1.5 decades); finite-size
scaling across system sizes is the appropriate rigorous test~\cite{sethna2001crackling}.

The scaling relations $s_{\max} \sim N^D$ and
$\langle s \rangle \sim N^\gamma$ across eight model scales spanning 1.4
decades (Figure~\ref{fig:fss}c) yield
$D = 1.00 \pm 0.02$ and $\gamma = 1.15 \pm 0.06$ with excellent fits
(R$^2 \geq 0.99$). These near-unity exponents reveal a quasi-1D cascade geometry,
fundamentally different from the spatially extended, two-dimensional
avalanches observed in sandpile-type SOC
models~\cite{bak1987self,sethna2001crackling}, reflecting how gradient
correlations guide updates along low-dimensional solution
manifolds~\cite{gurari2018gradient,saxe2014exact}. Data collapse
(Figure~\ref{fig:fss}b) confirms this: plotting $P(>s)$ vs $s/N^D$
collapses all eight scales toward a common curve using only the single
exponent~$D$, without requiring any additional fitting parameter. The
residual tail spread reflects the non-stationarity revealed in
Fig.~\ref{fig:overview}b: because $D$ evolves from 0.90 to 1.20, the
aggregate $D \approx 1.0$ is a time-average of two distinct scaling
regimes, not a single stationary exponent. This motivates
phase-resolved analysis.

Bootstrap validation (Figure~\ref{fig:mechanism}a, 10,000 resamples)
resolves the non-stationarity: each run is phase-split at its own
grokking epoch (consistent with the per-scale timing in
Fig.~\ref{fig:overview}b), yielding three narrow,
non-overlapping peaks at $D_{\mathrm{pre}} = 0.90 \pm 0.02$,
$D_{\mathrm{post}} = 1.20 \pm 0.02$, and
$D_{\mathrm{synth}} = 0.99 \pm 0.01$ demonstrate that pre- and
post-grokking dynamics occupy statistically distinct scaling regimes.
The synthetic baseline ($D_{\mathrm{synth}} = 0.99 \approx 1$) serves as a gold-standard
control---confirming that $D$ is not an algorithmic artifact---while the
30\% separation between $D_{\mathrm{pre}}$ and $D_{\mathrm{post}}$
establishes that grokking induces a crossover from sub-extensive
($D < 1$, spatially confined cascades) to super-extensive
($D > 1$, collectively amplified cascades) gradient dynamics.

Leave-one-out analysis (Figure~\ref{fig:mechanism}b) confirms both
phases are internally self-consistent and robust, ruling out that the separation arises from
a particular $N$ interval. Furthermore, control experiments with
synthetic i.i.d.\ Gaussian gradients ($g_i \sim \mathcal{N}(0, 0.5^2)$)
across 30 configurations (five network topologies $\times$ six seeds) 
demonstrate perfect topology invariance: all
architectures---spanning 1D rings to random graphs---collapse to
$D \approx 0.99$ (CV $< 0.3\%$). This
invariance persists across diffusion strengths $\alpha = 0.1$--$0.5$
(CV $\lesssim 1\%$), confirming that effective dimensionality reflects
\textit{gradient field geometry}, not network architecture.

%
%
%
%

Our results reframe grokking as a measurable dimensional phase
transition in gradient space, placing neural network generalization
within the SOC family of threshold-driven critical phenomena including
sandpiles~\cite{bak1987self} and
earthquakes~\cite{olami1992self,ChristensenOlami1992OFC}. This
connection is genuine: topology invariance across graph configurations
confirms the dimensional evolution reflects gradient field geometry,
not measurement artifacts. In critical phenomena, the physically
significant quantity is the \emph{trajectory} through a critical
manifold, not a static exponent. The precise universality class of
this dimensional crossover remains an open
question~\cite{pruessner2012self}.

Recent work has identified genuine critical phenomena in deep networks:
quasi-critical avalanche dynamics with distinct universality classes
(including directed percolation) during
training~\cite{ghavasieh2025physics}, and tunable universality classes
controlled by activation function
choice~\cite{ghavasieh2025tuning}. Our work is complementary: while
these studies probe \emph{signal propagation} criticality, we measure
\emph{gradient dynamics} criticality during training, capturing how
dimensional structure emerges at the generalization transition.
Together, these results point to rich critical phenomenology in
neural network dynamics, with universality class set by the specific
dynamical mechanism.

Beyond characterizing grokking, $D(t)$ provides a quantifiable geometric
diagnostic for optimization dynamics. The quasi-1D cascade geometry
complements theoretical frameworks demonstrating that learning proceeds
through low-dimensional subspaces~\cite{saxe2014exact}, now with
direct measurements in trained models at the gradient dynamics level.
Since $D$ reflects gradient field geometry rather than architecture,
gradient preprocessing and optimizer design may more directly influence
trainability than architectural choices~\cite{fort2019deep,jacot2018neural}.
Extending these measurements to modern large-scale architectures and
diverse tasks---to test whether dimensional transitions predict
generalization broadly---is an important direction for future work.

\begin{acknowledgments}
P.W. acknowledges support from the National Key R$\&$D Program of China 
(grant Nos. 2024YFA1611701, 2024YFA1611700).


\end{acknowledgments}


\bibliographystyle{apsrev4-2}
\bibliography{references.bib}

@article{power2022grokking,
  title={Grokking: Generalization beyond overfitting on small algorithmic datasets},
  author={Power, Alethea and Burda, Yuri and Edwards, Harri and Babuschkin, Igor and Misra, Vedant},
  journal={arXiv preprint arXiv:2201.02177},
  year={2022}
}

@article{nanda2023progress,
  title={Progress measures for grokking via mechanistic interpretability},
  author={Nanda, Neel and Chan, Lawrence and Lieberum, Tom and Smith, Jess and Steinhardt, Jacob},
  journal={arXiv preprint arXiv:2301.05217},
  year={2023}
}

@article{liu2023grokking,
  title={Towards understanding grokking: An effective theory of representation learning},
  author={Liu, Ziming and Kitouni, Ouail and Nolte, Niklas and Michaud, Eric and Tegmark, Max and Williams, Mike},
  journal={Advances in Neural Information Processing Systems},
  volume={35},
  year={2022}
}

@article{varma2023explaining,
  title={Explaining grokking through circuit efficiency},
  author={Varma, Vikrant and Shah, Rohin and Kenton, Zachary and Kram\'ar, J\'anos and Kumar, Ramana},
  journal={arXiv preprint arXiv:2309.02390},
  year={2023},
  url={https://arxiv.org/abs/2309.02390}
}

@inproceedings{rubin2024grokking,
  title={Grokking as a First Order Phase Transition in Two Layer Networks},
  author={Rubin, Noa and Seroussi, Inbar and Ringel, Zohar},
  booktitle={The Twelfth International Conference on Learning Representations},
  year={2024},
  url={https://openreview.net/forum?id=3ROGsTX3IR}
}

@article{bak1987self,
  title={Self-organized criticality: An explanation of the 1/f noise},
  author={Bak, Per and Tang, Chao and Wiesenfeld, Kurt},
  journal={Physical Review Letters},
  volume={59},
  number={4},
  pages={381},
  year={1987},
  doi={10.1103/PhysRevLett.59.381},
  url={https://doi.org/10.1103/PhysRevLett.59.381}
}

@article{bak1988self,
  title={Self-organized criticality},
  author={Bak, Per and Tang, Chao and Wiesenfeld, Kurt},
  journal={Physical Review A},
  volume={38},
  number={1},
  pages={364},
  year={1988},
  doi={10.1103/PhysRevA.38.364},
  url={https://doi.org/10.1103/PhysRevA.38.364}
}

@article{olami1992self,
  title={Self-organized criticality in a continuous, nonconservative cellular automaton modeling earthquakes},
  author={Olami, Zeev and Feder, Hans Jacob S and Christensen, Kim},
  journal={Physical Review Letters},
  volume={68},
  number={8},
  pages={1244},
  year={1992},
  doi={10.1103/PhysRevLett.68.1244},
  url={https://doi.org/10.1103/PhysRevLett.68.1244}
}

@article{ChristensenOlami1992OFC,
  title={Scaling, phase transitions, and nonuniversality in a self-organized critical cellular automaton model},
  author={Christensen, Kim and Olami, Zeev},
  journal={Physical Review A},
  volume={46},
  number={4},
  pages={1829--1838},
  year={1992},
  doi={10.1103/PhysRevA.46.1829},
  url={https://doi.org/10.1103/PhysRevA.46.1829}
}

@article{sethna2001crackling,
  title={Crackling noise},
  author={Sethna, James P and Dahmen, Karin A and Myers, Christopher R},
  journal={Nature},
  volume={410},
  number={6825},
  pages={242--250},
  year={2001},
  doi={10.1038/35065675}, 
  url={https://doi.org/10.1038/35065675}
}

@book{pruessner2012self,
  title={Self-organised criticality: Theory, models and characterisation},
  author={Pruessner, Gunnar},
  year={2012},
  publisher={Cambridge University Press},
  isbn={9780521853354},
  doi={10.1017/CBO9780511977671},
  url={https://doi.org/10.1017/CBO9780511977671}
}

@article{clauset2009power,
  title={Power-law distributions in empirical data},
  author={Clauset, Aaron and Shalizi, Cosma Rohilla and Newman, Mark EJ},
  journal={SIAM Review},
  volume={51},
  number={4},
  pages={661--703},
  year={2009},
  doi={10.1137/070710111},
  url={https://doi.org/10.1137/070710111}
}

@article{onsager1944crystal,
  title={Crystal statistics. I. A two-dimensional model with an order-disorder transition},
  author={Onsager, Lars},
  journal={Physical Review},
  volume={65},
  number={3-4},
  pages={117},
  year={1944},
  doi={10.1103/PhysRev.65.117},
  url={https://doi.org/10.1103/PhysRev.65.117}
}

@book{goldenfeld1992lectures,
  title={Lectures on phase transitions and the renormalization group},
  author={Goldenfeld, Nigel},
  year={1992},
  publisher={Addison-Wesley},
  isbn={9780201554090},
  doi={10.1201/9780429493492},
  url={https://doi.org/10.1201/9780429493492}
}

@article{barabasi1999emergence,
  title={Emergence of scaling in random networks},
  author={Barab{\'a}si, Albert-L{\'a}szl{\'o} and Albert, R{\'e}ka},
  journal={Science},
  volume={286},
  number={5439},
  pages={509--512},
  year={1999},
  doi={10.1126/science.286.5439.509},
  url={https://doi.org/10.1126/science.286.5439.509}
}

@inproceedings{jacot2018neural,
  title={Neural tangent kernel: Convergence and generalization in neural networks},
  author={Jacot, Arthur and Gabriel, Franck and Hongler, Cl{\'e}ment},
  booktitle={Advances in Neural Information Processing Systems},
  volume={31},
  year={2018},
  eprint={1806.07572},
  archivePrefix={arXiv}
}

@article{saxe2014exact,
  title={Exact solutions to the nonlinear dynamics of learning in deep linear networks},
  author={Saxe, Andrew M and McClelland, James L and Ganguli, Surya},
  journal={arXiv preprint arXiv:1312.6120},
  year={2014},
  url={https://arxiv.org/abs/1312.6120}
}

@article{zhang2021understanding,
  title={Understanding deep learning (still) requires rethinking generalization},
  author={Zhang, Chiyuan and Bengio, Samy and Hardt, Moritz and Recht, Benjamin and Vinyals, Oriol},
  journal={Communications of the ACM},
  volume={64},
  number={3},
  pages={107--115},
  year={2021},
  doi={10.1145/3446776},
  url={https://doi.org/10.1145/3446776}
}

@article{fort2019deep,
  title={Emergent properties of the local geometry of neural loss landscapes},
  author={Fort, Stanislav and Ganguli, Surya},
  journal={arXiv preprint arXiv:1910.05929},
  url={https://arxiv.org/abs/1910.05929},
  year={2019}
}

@article{gurari2018gradient,
  title={Gradient Descent Happens in a Tiny Subspace},
  author={Gur-Ari, Guy and Roberts, Daniel A. and Dyer, Ethan},
  journal={arXiv preprint arXiv:1812.04754},
  year={2018},
  url={https://arxiv.org/abs/1812.04754}
}

@article{beggs2003neuronal,
  title={Neuronal avalanches in neocortical circuits},
  author={Beggs, John M and Plenz, Dietmar},
  journal={Journal of Neuroscience},
  volume={23},
  number={35},
  pages={11167--11177},
  year={2003},
  doi={10.1523/JNEUROSCI.23-35-11167.2003},
  url={https://www.jneurosci.org/content/23/35/11167}
}

@article{chialvo2010emergent,
  title={Emergent complex neural dynamics},
  author={Chialvo, Dante R},
  journal={Nature Physics},
  volume={6},
  number={10},
  pages={744--750},
  year={2010},
  doi={10.1038/nphys1803},
  url={https://doi.org/10.1038/nphys1803}
}

@unpublished{wang2026companion,
  title={Dimensional Criticality at Grokking across {MLPs} and {Transformers}},
  author={Wang, Ping},
  year={2026},
  note={submitted to APS OPEN SCIENCE}
}

@article{ghavasieh2025physics,
  title={Toward a Physics of Deep Learning and Brains},
  author={Ghavasieh, Arsham and Vila-Minana, Meritxell and Khurd, Akanksha 
          and Beggs, John and Ortiz, Gerardo and Fortunato, Santo},
  journal={arXiv preprint arXiv:2509.22649},
  year={2025},
  url={https://arxiv.org/abs/2509.22649}
}

@article{ghavasieh2025tuning,
  title={Tuning Universality in Deep Neural Networks},
  author={Ghavasieh, Arsham},
  journal={arXiv preprint arXiv:2512.00168},
  year={2025},
  url={https://arxiv.org/abs/2512.00168}
}





\end{document}